%% file: 0-main.tex
\documentclass{article}

\usepackage{microtype}
\usepackage{graphicx}
\usepackage{subfigure}
\usepackage{booktabs} % for professional tables

\usepackage{hyperref}

\usepackage[accepted]{icml2023}

\usepackage{amsmath}
\usepackage{amssymb}
\usepackage{mathtools}
\usepackage{amsthm}
\usepackage{bbold} %TODO

\usepackage[capitalize,noabbrev]{cleveref}

\theoremstyle{plain}

\theoremstyle{definition}

\theoremstyle{remark}

\usepackage[textsize=tiny]{todonotes}

\icmltitlerunning{Diagnosis, Feedback, Adaptation: A Human-in-the-Loop Framework for Test-Time Policy Adaptation}

\begin{document}

\twocolumn[
\icmltitle{Diagnosis, Feedback, Adaptation: \\A Human-in-the-Loop Framework for Test-Time Policy Adaptation}




\begin{icmlauthorlist}
\icmlauthor{Andi Peng}{mit}
\icmlauthor{Aviv Netanyahu}{mit}
\icmlauthor{Mark Ho}{nyu}
\icmlauthor{Tianmin Shu}{mit}
\icmlauthor{Andreea Bobu}{berkeley}
\icmlauthor{Julie Shah}{mit}
\icmlauthor{Pulkit Agrawal}{mit}
\end{icmlauthorlist}

\icmlaffiliation{mit}{Massachusetts Institute of Technology}
\icmlaffiliation{nyu}{New York University}
\icmlaffiliation{berkeley}{University of California, Berkeley}

\icmlcorrespondingauthor{Andi Peng}{andipeng@mit.edu}

\icmlkeywords{Machine Learning, ICML}

\vskip 0.3in
]



\printAffiliationsAndNotice{} % otherwise use the standard text.

\begin{abstract}
Policies often fail due to distribution shift---changes in the state and reward that occur when a policy is deployed in new environments. Data augmentation can increase robustness by making the model invariant to \textit{task-irrelevant} changes in the agent's observation.
However, designers don't know which concepts are irrelevant \textit{a priori}, especially when different end users have different preferences about how the task is performed.  
We propose an interactive framework to leverage feedback directly from the user to identify \textit{personalized} task-irrelevant concepts. Our key idea is to generate \textit{counterfactual demonstrations} that allow users to quickly identify possible task-relevant and irrelevant concepts. The knowledge of task-irrelevant concepts is then used to perform data augmentation and thus obtain a policy adapted to personalized user objectives. 
We present experiments validating our framework on discrete and continuous control tasks with real human users. Our method (1) enables users to better understand agent failure, (2) reduces the number of demonstrations required for fine-tuning, and (3) aligns the agent to individual user task preferences.
\end{abstract}

\input{1-intro.tex}
\input{2-formulation.tex}
\input{3-approach.tex}
\input{4-experiments.tex}
\input{5-results.tex}
\input{6-related_work}
\input{7-discussion}
\input{8-acknowledgements}

\bibliography{references}
\bibliographystyle{icml2023}

%%%%%%%%%%%%%%%%%%%%%%%%%%%%%%%%%%%%%%%%%%%%%%%%%%%%%%%%%%%%%%%%%%%%%%%%%%%%%%%
%%%%%%%%%%%%%%%%%%%%%%%%%%%%%%%%%%%%%%%%%%%%%%%%%%%%%%%%%%%%%%%%%%%%%%%%%%%%%%%
% APPENDIX
%%%%%%%%%%%%%%%%%%%%%%%%%%%%%%%%%%%%%%%%%%%%%%%%%%%%%%%%%%%%%%%%%%%%%%%%%%%%%%%
%%%%%%%%%%%%%%%%%%%%%%%%%%%%%%%%%%%%%%%%%%%%%%%%%%%%%%%%%%%%%%%%%%%%%%%%%%%%%%%
\newpage
\input{9-appendix}

\end{document}

%% file: 1-intro.tex
\section{Introduction}
\label{sec:intro}

\begin{figure}[ht]
    \centering
    \includegraphics[width=.5\textwidth]{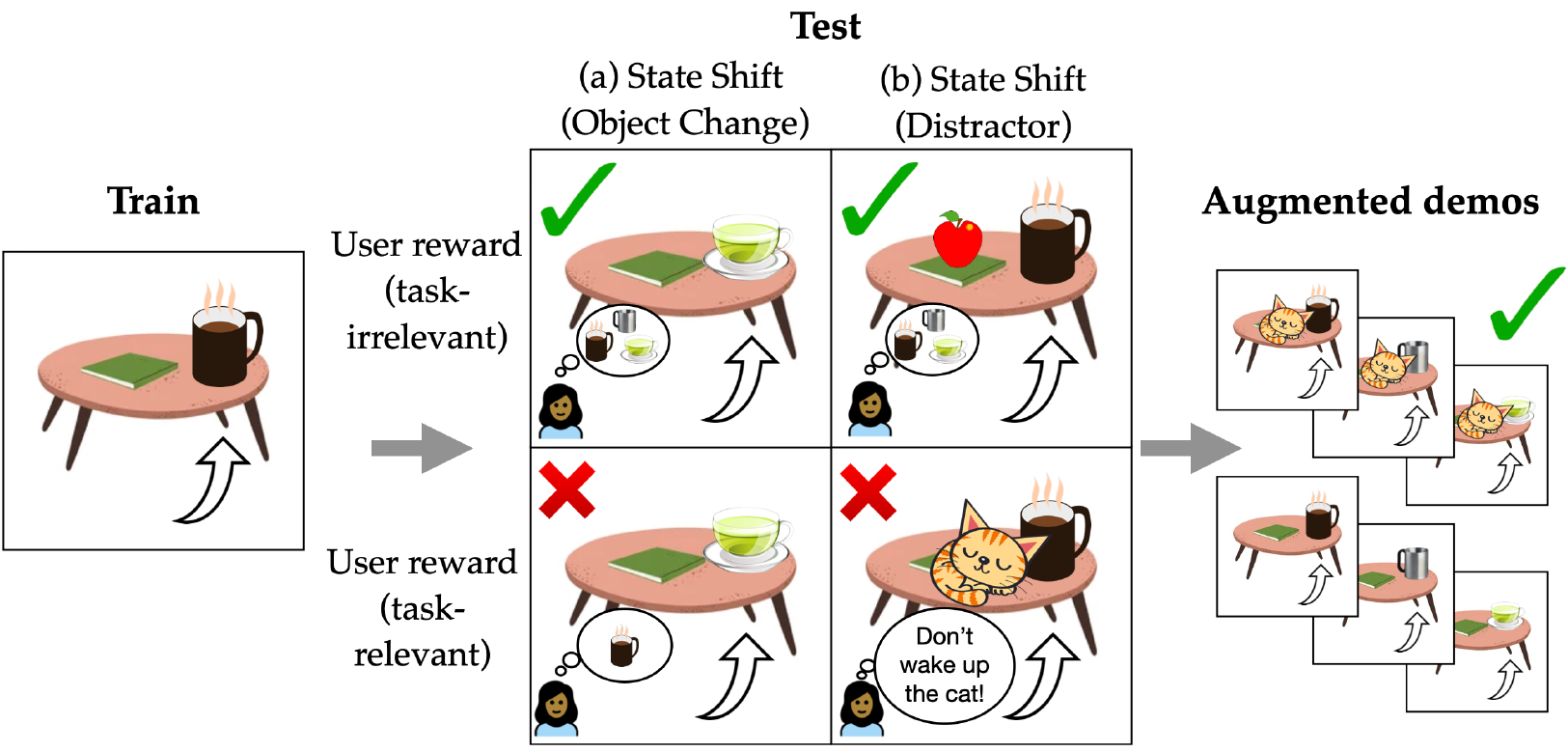}
    \caption{Illustrative distribution shifts for task: \textbf{``Get my mug."} Shifted concepts can be state-based (a changed object) and also reward-based (dependent on user preference). We can deploy data augmentation for \textit{task-irrelevant} shifts (green checks), a subset where the modified state \textit{does not impact} desired policy behaviour.}
    \label{fig:shifts}
\end{figure}

Consider a user who purchases a factory-trained robot for use in their home (Figure 1). They task the robot: ``get my mug'' and are confused when it unexpectedly crashes. Despite the factory's best attempts at teaching the robot common visual concepts like mug \textit{shape} and \textit{material}, the robot may face a distribution shift from the observations it saw during training, e.g. the robot only saw ceramic and metal mugs in the factory but not the user's glass mug.

\begin{figure*}
    \centering
    \includegraphics[width=.9\textwidth]{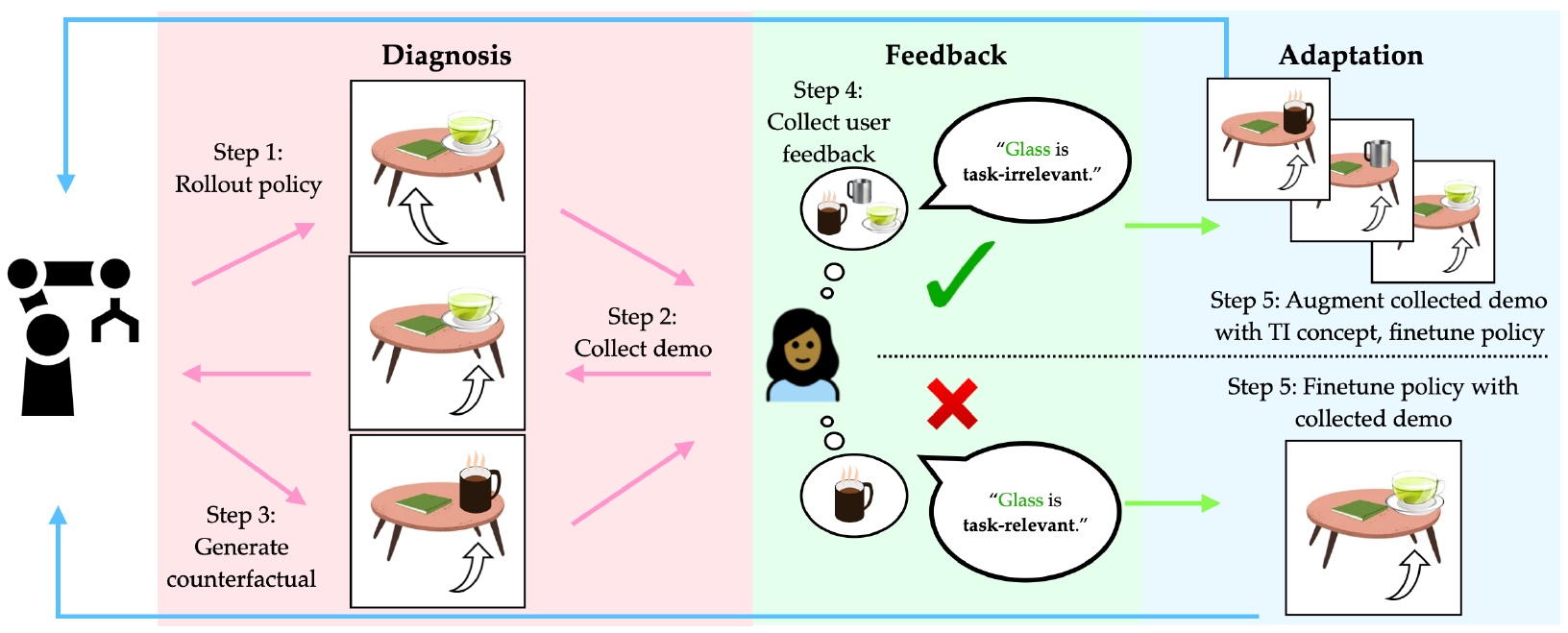}
    \caption{Overview of our framework. At test time, we perform a three-phrase procedure to (1) generate counterfactual demonstrations to help the user diagnose shifted state concepts, (2) collect user feedback on which visual concepts are task-irrelevant (TI) vs. task-relevant (TR), and then (3) augment the collected demonstration based on invariant visual concepts identified by the user to finetune the policy.}
    \label{fig:pipeline}
\end{figure*}

What can we, as designers, do to fix the robot? We could retrain the robot via reinforcement learning, but such training is expensive and requires the user to be an expert at reward design \cite{amodei2016concrete}. Alternatively, we could ask the user for diverse demonstrations of them picking up the mug for finetuning the robot via imitation learning~\cite{schaal1999imitation}, but doing so would be time-consuming (and unscalable) for the user. Fortunately, we can teach the robot more efficiently via data augmentation \cite{baran2019safe,tobin2017domain}, a technique that varies, or augments, concepts in the state that we know to not impact the task. For example, if the user intends for all mugs to be similarly handled (and the glass mug has a similar size and shape to already seen ceramic mugs), we could ``recycle'' actions the robot knows for picking up ceramic mugs to also teach it to pick up the glass mug. But what if the user actually considers the glass object to be a cup (instead of a mug), i.e. their \textit{reward} implicitly includes the material of the object being important to the task? Unfortunately, we don't know this ahead of time, i.e. knowing whether the shifted concept is \textit{task-irrelevant} (TI) or \textit{task-relevant} (TR) for the desired task expressed by a user is impossible \textit{a priori}.

Our insight is that \textit{end users are uniquely positioned to recognize which concepts are relevant or irrelevant for their desired task}. If we had a way to reliably and quickly query the human for irrelevant concepts adversely impacting the robot's behaviour, then we could leverage data augmentation for those concepts and significantly reduce the number of additional demonstrations required for finetuning the robot.

How do we elicit this feedback? Problematically, humans are not adept at identifying feature-specific causes of black box model failures \cite{doshi2017towards,goyal2019counterfactual}. For example, if the robot from above crashed into the coffee table, it would be difficult for the user to identify that the mug's material being glass was the true source of failure vs. other visual concepts such as the coffee table or location of the mug. Ergo, inspired by the interpretability literature, we propose a \textit{counterfactual} approach to identify failure \cite{olson2019counterfactual}. Consider that the human also observes a trajectory of the robot successfully retrieving the mug in the same scenario but with a single change -- the \textit{mug being ceramic instead of glass}. Being able to contrast the two trajectories of successful and unsuccessful behaviour can better position the user to identify visual concepts impacting failure. Here, the user may identify mug material to be a task-irrelevant concept, which can then be used to make the policy invariant to mug-material via data augmentation.% (assuming knowledge of all possible mug-materials) and without collecting additional user demonstrations. 
%where the robot communicates a \textit{contrastive demonstration} showing what visual concepts (or features) in the observation would have needed to change for the policy to have produced different behaviour. 

The main challenge in implementing this idea is that the robot cannot provide a successful demonstration in the new scenario where it has failed. Thus, we ask the user for a single demonstration.
To generate a counterfactual demonstration, we ask which visual concept, if changed, would cause the policy to output actions that match those of the user's. For this, we search through the space of known visual concepts to create visual observations using a generative model for which the policy's actions matches the demonstration. 
%that leads the using a generative model  for generating many visual observations corresponding to different instantiations of the visual concept until we find
%generate visual observations corresponding to different instantiations of the visual concept until we find the one that makes the policy actions match the demonstration (i.e., finding a successful robot trajectory). 
%then performing a search through the visual concept space for generating observations that would make the policy output actions similar to that in the demonstration.
%similar actions for the policy had those concepts been changed. 
In some cases, there may not be a unique visual concept that can be adjusted to obtain the desired behaviour. Thus, we provide the user with the unsuccessful and successful trajectories along with the identified visual concepts for the user to verify if the inference was correct.
%different visual concepts might lead to the same Finally we provide the user with the two trajectories and the visual concept 
%Through more accurate user feedback of irrelevant concepts, we can then perform better data augmentation of those concepts for policy finetuning with less user effort.
%This feedback improves the accuracy of user identification of task-irrelevant visual concepts.
%more accurate user feedback of irrelevant concepts, we can then perform better data augmentation of those concepts for policy finetuning with less user effort.

We call our full framework DFA (Diagnosis, Feedback, Adaptation) (Figure 2). In the \textit{diagnosis phase}, we show a failed trajectory to the user, e.g. the robot crashing into the coffee table, and collect a demonstration of the user's desired behaviour, e.g. successful retrieval of the glass mug. We then use this demonstration to generate counterfactual trajectories by modifying visual concepts (i.e., what actions would the policy output if mug material was changed) until we find a trajectory that matches the demonstration. 
%that would have produced similar actions for the policy \textit{had specific visual concepts}, e.g. mug material, \textit{of the state changed}. 
Our main hypothesis is that not only can counterfactual trajectories aid the identification of concepts leading to robot failure, e.g. the mug being glass, but they can also be leveraged to query for human feedback regarding whether or not the robot should be invariant to these concepts for the user's desired task. We do this in the \textit{feedback phase}, where the user verifies the counterfactuals to confirm whether the identified concept is task-irrelevant to the user's reward, e.g. \textit{should the mug being made of glass impact the task}? To close the loop, we leverage these identified irrelevant concepts in the \textit{adaptation phase} to perform efficient finetuning via augmentation of the user-identified task-irrelevant concepts.

We test our framework in three domains consisting of both discrete and continuous control tasks with real human users. Through human experiments, we verify our main hypothesis that user feedback resulting from counterfactual demonstrations significantly improves the accuracy of user-identified TI concepts as well as the data efficiency of policy finetuning with \textit{less user effort}. These findings illustrate a promising direction into leveraging end users to more efficiently perform interactive alignment of robotic policies at test-time.

%% file: 2-formulation.tex
% \section{Inferring and fixing distribution shift}
\section{Problem Setup}
\label{sec:formulation}
We consider the problem of efficiently adapting a model-free policy to test scenarios under distribution shift. We define success as \textit{achieving high rewards on the test scenarios after policy finetuning} while \textit{minimizing} the number of expert demonstrations required from the user. We consider test scenarios where potentially both the state and user-desired reward may have shifted from training. 

\textbf{Preliminaries.} 
We consider environments represented by a tuple $M = \langle \mathcal{S}, \mathcal{A}, \mathcal{T}, \mathcal{R} \rangle$, where $\mathcal{S}$ is the state space, $\mathcal{A}$ the action space, $\mathcal{T}:\mathcal{S} \times \mathcal{A} \times \mathcal{S} \rightarrow [0,1]$ the transition probability distribution, and $\mathcal{R}:\mathcal{S} \times \mathcal{A} \rightarrow \mathbb{R}$ the reward function.
A parameterized policy is denoted as $\pi_\theta:\mathcal{S}\rightarrow \mathcal{A}$, which is trained end-to-end  via any method such as imitation learning. 
In this work, we train agents from raw pixel inputs and thus represent observations as RGB images.
%such as for visual navigation or camera-based high-level manipulation and rearrangement tasks. We therefore represent observations as RGB images.

% We now formalize the \textbf{distribution shift inference paradigm} faced by users when inferring concept-level failures of policies deployed in new environments.

\textbf{Distribution shift.} Suppose policy $\pi_{\theta}$ is trained on initial states $S_0^{\mathrm{tr}}\subseteq\mathcal{S}$ with reward $\mathcal{R}$ and deployed on initial test states $S_0^{\mathrm{te}}\subseteq\mathcal{S}$ with reward $\mathcal{R}'$. End-to-end policies suffer from the problem of distribution shift \cite{shimodaira2000improving}, i.e. $\pi_\theta$ can behave arbitrarily if $S_0^{\mathrm{te}}$ changes from $S_0^{\mathrm{tr}}$. Moreover, $\mathcal{R}$ often under-specifies the true reward \cite{agrawal2022task}, and so the user's intended $\mathcal{R}'$ may also implicitly change from the $\mathcal{R}$ the policy was trained on. We refer to these two shift types as \textit{state shifts} and \textit{reward shifts}.

\textbf{State shifts.}
To ground the potential change from $S_0^{\mathrm{tr}}$ to $S_0^{\mathrm{te}}$, we define visual concepts of objects present in $\mathcal{S}$ as $\Phi:\mathcal{S}\to\mathcal{S}_{\Phi}$, i.e. \textit{state abstractions} that contain object-centric concepts, such as color, relevant to the task. While we assume \textit{a priori} knowledge of the family of concepts that can vary at test-time during training, $\Phi(S_0^{\mathrm{tr}})$, we do not assume any knowledge of their test-time instantiation $\Phi(S_0^{\mathrm{te}})$, e.g. color being \textit{blue} may not be in the training set. As illustrated in Figure~\ref{fig:shifts}, potential shifts in $\mathcal{S}_{\Phi}$ can be due to 
new instantiations of object concepts (e.g. color changes from \textit{red} to \textit{orange}), or the introduction of distractors (e.g. a cat appears). We define these shifts as transformations of initial train states in $\Phi(S_0^{\mathrm{tr}})$ into initial test states in $\Phi(S_0^{\mathrm{te}})$, which are \textbf{unknown} to both the model and end user. %For ease of exposition, we consider only the initial state as exhibiting full information regarding the shift across a trajectory, although we can extend the formulation to encompass full state sequences.
%Let the set $\phi_{\mathrm{shift}}$ denote the concept(s) $\{\phi^i_j\_{i,j}$ shifting between $S$ and $S'$ in the case of shifted object visual concepts or new distractors that are represented in $\mathcal{S}_{\Phi}$.

\textbf{Assumption 1:}
Although we do not assume $\mathcal{S}_{\Phi}$ comprehensively captures all possible state concepts impacting behaviour, we assume it captures a subset.
% There may exist additional visual concepts not included in $\mathcal{S}_{\Phi}$ that are relevant to the task, e.g. position and orientation of the object. We \textbf{do not} assume that the visual concept instantiations represented by $\Phi(s)$ are \textit{comprehensive} of all visual concepts relevant for planning.

This assumption is reasonable for tasks such as navigation or high-level object manipulation, where common state shifts may be object-centric (e.g. geometric or visual object properties like shape or color may be the reason a policy fails). 

\textbf{Reward shifts.}
Although policy $\pi_{\theta}$ is trained with reward function $\mathcal{R}$ in the training environment, at test time, a user may want the agent to optimize a different reward function, $\mathcal{R}'$. For example, Figure~\ref{fig:shifts} illustrates that although the task is specified as ``get the fruit,'' the policy may have been trained only for $\mathcal{R}$: ``get the apples,'' whereas a user may instead prefer $\mathcal{R}'$: ``get \textit{any} fruit.'' Therefore, whether a shift in the abstracted state space $\mathcal{S}_{\Phi}$ is task-irrelevant (TI) or not is also conditioned on the end user's reward. Let $\mathcal{R}'$ be the user desired reward, which may or may not be the reward, $\mathcal{R}$, used for the policy training. This potential shift of reward is also \textbf{unknown} to both the model and end user.

%% file: 3-approach.tex
\section{Our Approach: DFA}
\label{sec:approach}

We describe our approach for leveraging user feedback to efficiently perform policy adaptation. To do so, we must possess (1) a state editor capable of making concept-level state changes and (2) a way to query the end user for TI concepts that are grounded in their $R'$.

\begin{figure}[t]
    \centering
    \includegraphics[width=.35\textwidth]{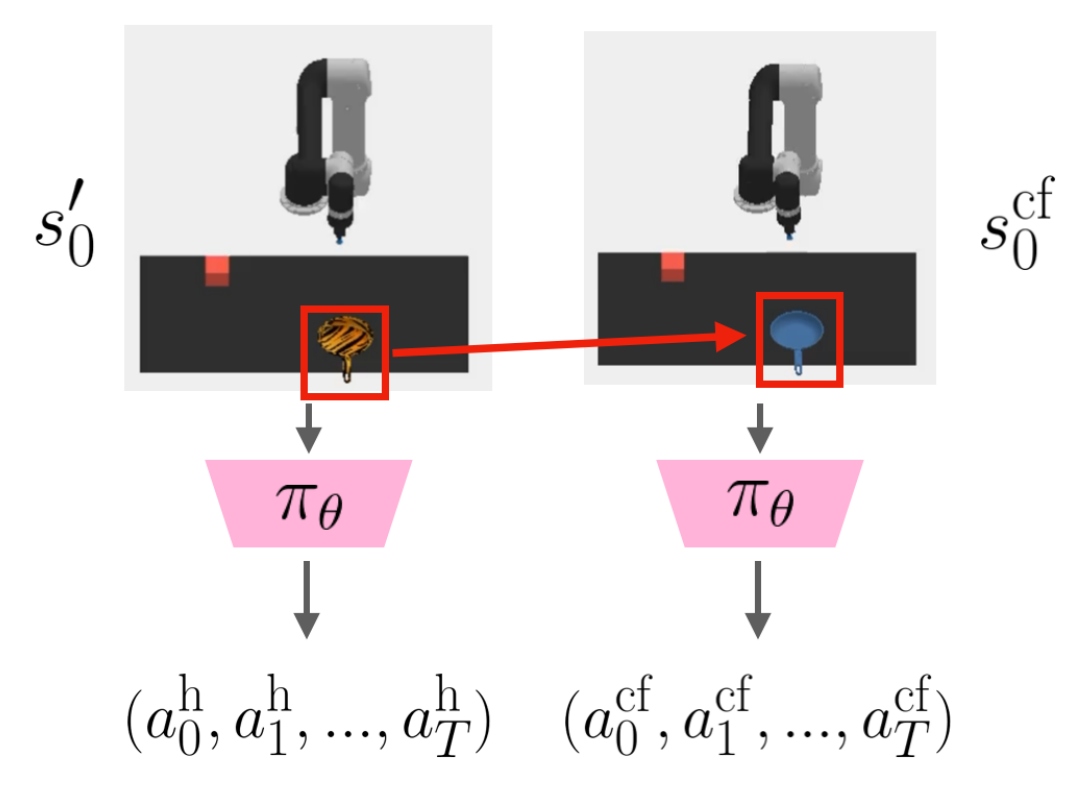}
    \caption{Our approach generates counterfactual demonstrations (right) for a given human demonstration (left) --- explaining what visual concept(s) of the test state needed to have changed such that the policy produced actions similar to the user desired actions.}
    \label{fig:cf}
\end{figure}

\textbf{Mapping observations to visual concepts.}
We begin by defining the state editor. Let $\Phi:\mathcal{S}\to\mathcal{S}_{\Phi}$ be a mapping from the observation (image) to a vector describing its object-based visual concepts. We denote $\Phi(s)=(\phi_0^0, \phi_0^1,\dots,\phi_0^k,\dots,\phi_n^0, \phi_n^1,\dots,\phi_n^k)$
% Let states in $S$ be parameterized via object-based visual concepts $\mathcal{S}_{\Phi}=\{(\phi_0^0, \phi_0^1,\dots,\phi_0^k,\dots,\phi_n^0, \phi_n^1,\dots,\phi_n^k)\}_{s\in\mathcal{S}}$, 
where $\phi_i^j\in\{0,1\}^m$ indicates whether object $i$ is present in state $s$ (e.g., a fruit) and its object-specific concept $j$ (e.g., \textit{color}). $\phi_i^j$ is a one-hot encoding representing the object-specific concept \textit{instantiation} (e.g. \textit{red}).
% The instantiation of this concept (e.g. the fruit color is \textit{red}) is discrete and expressed by a one-hot encoding value of $\phi_i^j$.
% We denote the mapping from states to concept parameterization as $\Phi:\mathcal{S}\to\mathcal{S}_{\Phi}$.
 % We assume access to an ordering in concept space based on object IDs.
% Since $\mathcal{S}_{\Phi}$ may not capture all visual concepts present in an image state, mapping visual concepts back to states is conditioned on the state.
We assume access to $\Phi$ and its conditional inverse $\Phi^{-1}:\mathcal{S}_{\Phi}\times\mathcal{S}\to\mathcal{S}$, i.e. a visual generative model.
% There may exist additional visual concepts not included in $\mathcal{S}_{\Phi}$ that are relevant to the task, e.g. position and orientation of the object. We \textbf{do not} assume that the visual concept instantiations represented by $\Phi(s)$ are \textit{comprehensive} of all visual concepts relevant for planning.

\textbf{Assumption 2:} Let $\Delta\mathcal{S}_{\Phi}:=\{\Phi(s_1)-\Phi(s_2)|s_1,s_2\in\mathcal{S}\}$. We assume we possess a transformation function, i.e. image editor, $f:\mathcal{S}_{\Phi}\times\Delta\mathcal{S}_{\Phi}\to\mathcal{S}_{\Phi}$ capable of modifying visual concept representations. These changes may include editing existing concepts (turning a \textit{red} object \textit{orange}), removing objects entirely (by setting their value to $\Vec{0}$), or spawning new objects (by setting their value to $\neq\Vec{0}$).
We assume we know the space of edits $\Delta\phi$. We can modify visual concepts in state $s$ and generate state $s'$ as follows:
% $\Phi^{-1}\circ f \circ\Phi:\mathcal{S}\to\mathcal{S}$
$s'=\Phi^{-1}(f(\Phi(s),\Delta\phi),s)$ where $\Delta\phi\in\Delta\mathcal{S}_{\Phi}$.

These assumptions are reasonable due to 1) increasingly higher-fidelity robotic simulators for common attributes found in object-centric tasks (such as object shape and color) \cite{mittal2023orbit,li2022behavior} and 2) promising advancements in text-conditioned image editing \cite{gal2022image,kawar2022imagic}. Both approaches hold promise for creating and modifying realistic scenes and also require an abstracted state space grounded in a simulator.

We now demonstrate how we use $\Phi, \Phi^{-1}$ and $f$ to create counterfactual demonstrations for querying TI concepts from the user and perform efficient policy adaptation.

\textbf{Diagnosis: Generate counterfactuals}\\
We first sample an initial test state from a distribution over initial test states $s_0'\sim\mathcal{D}(S_0^{\mathrm{te}})$, and deploy policy $\pi_{\theta}$ to collect trajectory $\tau'=(s_0',a_0',\dots,s_T',a_T')$ (Algorithm~\ref{alg:algorithm} line 3).
% If $s_0'$ is out of distribution, $\tau'$ will result in failure. 
Since we do not know the true user-desired $R'$, we must query the human to determine whether the policy resulted in success or failure (Algorithm~\ref{alg:algorithm} line 4).
If the latter, we request one demonstration from the human $\tau_h=(s_0',a_0^h,\dots,s_T^h,a_T^h)$ (Algorithm~\ref{alg:algorithm} line 5).
% Let $\Phi(s_0')=(\phi_0^0,\phi_0^1, \dots, \phi_n^0,\phi_n^1)$.
We apply our abstract state editor $f$ to replace each concept $\phi_i^j\in\Phi(s_0')$ (e.g. color and shape) with alternative instantiations (e.g. \textit{red}, \textit{orange}, \textit{does not exist} etc.) and represent the changes made to a state as $\Delta\phi$.
These edits represent a search over the visual concept space that can possibly invoke contrastive behaviour from the policy resulting in a successful trajectory.
We do not assume access to the different concept instantiations used for training the policy, i.e. we treat the policy as a black-box. 
%that we are unable to obtain privileged access to.

The counterfactual state is then obtained from the edited visual concept representation via $\Phi^{-1}$. Altogether: 
$s_0^{\mathrm{cf}}=\Phi^{-1}( f(\Phi(s_0'),\Delta\phi),s_0')$.
Deploying $\pi_{\theta}$ on the counterfactual initial state results in trajectory $\tau_{cf}=(s_0^{\mathrm{cf}},a_0^{\mathrm{cf}},\dots,s_T^{\mathrm{cf}},a_T^{\mathrm{cf}})$ (Algorithm~\ref{alg:algorithm} line 7). 
Inspired by the \textit{minimum edit counterfactual problem} \cite{goyal2019counterfactual}, where counterfactuals for image classification are produced by searching through the feature space for the minimal modification required to change the model's classification output, we formulate a parallel for sequential decision-making by selecting the minimal concept change that results in actions close to the human demonstration: 

\begin{equation} 
\begin{aligned}
\underset{\Delta\phi}{\mathrm{argmin}}\sum_{i,j}\mathbb{1}_{\Phi(s_0')_i^j\neq\Phi(s_0^{\mathrm{cf}})_i^j}\\
    % \underset{\mathcal{I}=\{(i,j)|\Phi(s_0')_i^j\neq\Phi(s_0^{\mathrm{cf}})_i^j\}}{\mathrm{argmin}}|\mathcal{I}|\\    %\sum_{i,j}\mathbb{1}_{\Phi(s_0')_i^j\neq\Phi(s_0^{\mathrm{cf}})_i^j}\\
    \mathrm{s.t.}\, \forall t\in [T]:|a_t^{\mathrm{cf}}-a_t^h|<\epsilon
\end{aligned}
% $\mathrm{min}\sum_{i,j}\mathds{1}_{\phi_i^j\neq(\phi ')_i^j}$ where $s'=s(\Phi ')$.
\end{equation}

This can be intuitively thought of as minimizing the number of ``edits'' performed by state editor $f$ such that the policy produces actions similar to the demonstrated actions. We choose $L_1$ distance as a natural and general distance metric, although alternate metrics can be used, including those that compute distances between full trajectories \cite{berndt1994using}. Although we conduct exhaustive search in this work, alternate formulations could perform continuous relaxations that allow the objective to be directly optimizable via gradient descent depending on the assumed training distribution \cite{goyal2019counterfactual}.

% \textbf{Hypothesis 1:}
% \emph{Counterfactual demonstrations} will enable more accurate human inference of shifted state visual concepts $\phi_{\mathrm{shift}}$ compared to test behaviour alone.

\textbf{Feedback: Query TI concepts and augment data}\\
With these counterfactuals, we now have a human-interpretable manner of communicating visual concepts that enables user-guided data augmentation of TI concepts on their desired reward $\mathcal{R}'$. While our generated counterfactual contains information regarding the policy's hypothesis of what shifted concepts $\Delta\phi$ are, we do not know if these concepts are TI to the user desired reward $\mathcal{R}'$. Therefore, we query for user feedback regarding whether the identified $\Delta\phi$ is TI or task-relevant (TR), i.e. affects the desired behaviour, on reward $\mathcal{R}'$. For example, if the user had identified mug texture as the shifted visual concept causing failure, we then ask whether this concept is task-relevant or -irrelevant.

\textbf{Adaptation: Finetune policy}\\
Lastly, using the previously collected human demonstration, we perform efficient data augmentation of state sequences via modifying all instantiations of the identified TI concept(s) with state editor $f$. We then use these augmented demonstrations for finetuning our policy via supervised learning. The full experimental framework is below.

\begin{algorithm}[t!]
    \small
    \caption{: Fast adaptation with counterfactuals}
      	\begin{algorithmic}[1]
      	\STATE \textbf{Given}: policy $\pi_{\theta}$, test state $s_0'$, state editor $f$, human user
            \STATE \textcolor{orange}{/\ /\ Diagnosis}
      	\STATE Rollout $\pi_{\theta}$ on $s_0'$ for user
            \WHILE{user indicated failure}
                \STATE Collect user demo
                $\tau_\mathrm{h}: (s_0', a_0^h,...,s_T^h, a_T^h)$
                \STATE Initialize $D_{\mathrm{finetune}}: \{\tau_\mathrm{h}\}$
                \STATE Generate counterfactual
                $\tau_{\mathrm{cf}}: (s_0^{\mathrm{cf}}, a_0^{\mathrm{cf}},...,s_T^{\mathrm{cf}}, a_T^{\mathrm{cf}})$
                \STATE \textcolor{orange}{/\ /\ Feedback}
                \IF{$\tau_{\mathrm{cf}}$ actions close to $\tau_\mathrm{h}$}
                    \STATE Collect user feedback on $\Delta\phi$ and \\whether it is \textit{task-irrelevant}
              	\IF{YES}
                        \FOR{concept instantiation $\in[m]$ of $\Delta\phi$}
                            \STATE Create augmented demo $\tau_{\mathrm{aug}}:(\Tilde{s_0}, a_0^h,...\Tilde{s_T}, a_T^h)$
                            \STATE Update $D_{\mathrm{finetune}}\leftarrow D_{\mathrm{finetune}}\cup\{\tau_{\mathrm{aug}}\}$
                        \ENDFOR
                        %\IF{access to pretraining demos $D_{\mathrm{pretrain}}$}
                        %    \STATE Repeat steps 11-13 for all demos in $D_{\mathrm{pretrain}}$
                        %\ENDIF
                    \ENDIF
                \ENDIF
            \STATE \textcolor{orange}{/\ /\ Adaptation}
            \STATE Finetune $\pi_{\theta}$ using $D_{\mathrm{finetune}}$
            \ENDWHILE
      	\end{algorithmic}
\label{alg:algorithm}
\end{algorithm}

%While we do not assume access to the original training demonstrations, one could  perform the augmentation from Algorithm~\ref{alg:algorithm} lines 12-14 additionally on those demonstrations. We treat the policy as a black-box given to us at deployment and therefore do not perform this augmentation.

% \textbf{Hypothesis 2:}
% Counterfactuals for querying user feedback will enable efficient data augmentation and therefore user-desired policy performance with fewer demonstrations.

%% file: 4-experiments.tex
\section{Experimental domains and task generation}
\label{sec:experiments}

\begin{figure}
    \centering
    \includegraphics[width=.45\textwidth]{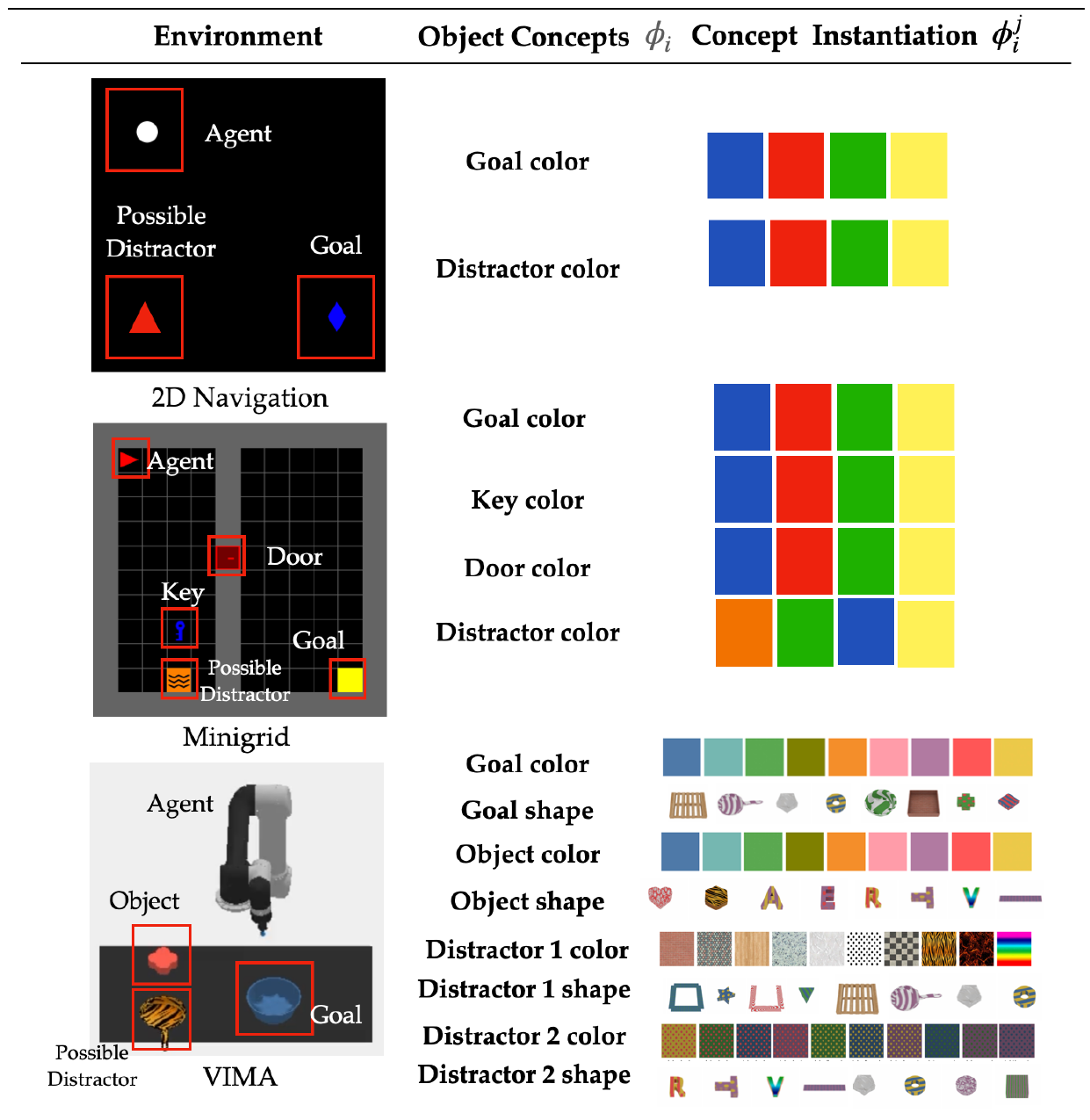}
    \caption{Environments and their possible concept instantiations. Note, VIMA contains more instantiations than depicted (there are 29 possible shapes and 81 possible colors in total).}
    \label{fig:features}
\end{figure}

We evaluate our full experimental pipeline on three environments consisting of both discrete and continuous control tasks: \textbf{2D Navigation}, \textbf{Minigrid}, and \textbf{VIMA}. Figure~\ref{fig:features} details state-editable visual concept instantiations $\phi^i_j$. For implementation details, please see Appendix~\ref{supp:implementation-details}.

\begin{figure*}
    \centering
    \includegraphics[width=1.\textwidth]{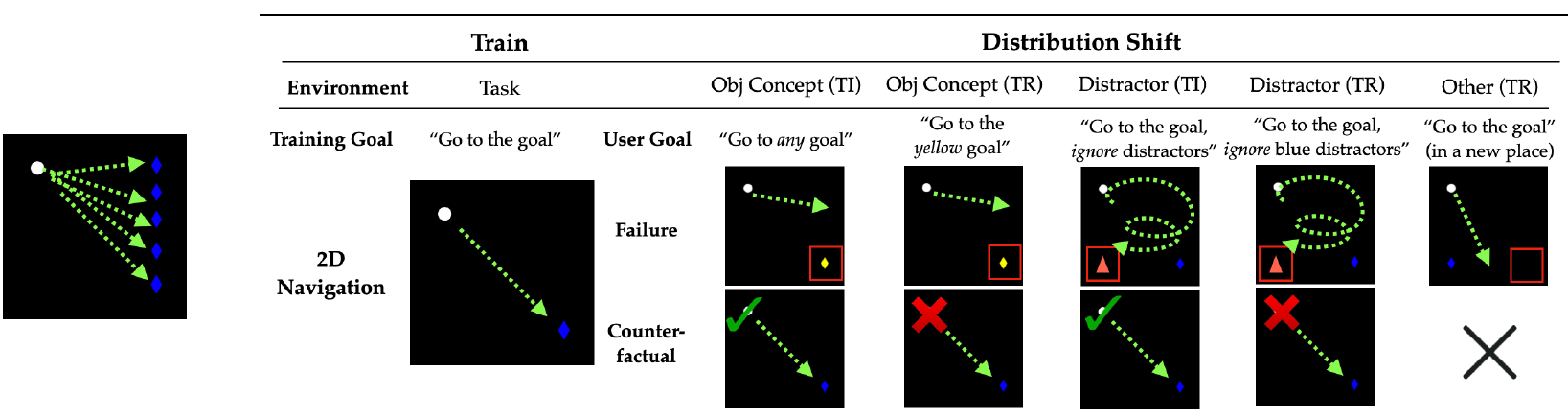}
    \caption{Example train task, possible test tasks, and their corresponding counterfactuals for each distribution shift on 2D Navigation.}
    \label{fig:envs-2d}
\end{figure*}

\textbf{2D Navigation.} To highlight an illustrative example (Figure \ref{fig:envs-2d}), we create a simple visual navigation environment where an agent is tasked with navigating to a goal object of one color while potentially faced with a distractor of a different color. Success is defined as being within radius $\epsilon$ of the final goal after 20 timesteps.
% States are fully-observable RGB images of dimension 36$\times$36$\times$3. The action space is continuous and represents the $(x,y)$ distance that the agent can move in 2D space. The environment was created to test a simple visual domain while preserving a continuous action space.

% There are two possible objects: goal and distractor. There is one visual concept (color) with 4 instantiations (red, green, blue, yellow).

\textit{Training task.} 
We generate a task, defined as ``go to the $<$\texttt{goal}$>$'', with an agent, a randomly sampled goal color, and no distractor. We place the goal in the bottom right corner of the grid and the agent (always white) in the top left corner. The train reward $\mathcal{R}$ is the agent's distance from the goal. We then create 10 demonstrations of length 20 by taking continuous actions from the agent's starting location to the goal object. We use these to train policy $\pi_{\theta}$ via supervised learning. Refer to Figure~\ref{fig:envs-2d} for an example.

\textit{Test tasks.}
We sample a test task for each of the 5 potential distribution shifts: shifted object concept (TI and TR), new distractor (TI and TR), and other (always TR). For the first and second shift type (modified object concept), we sample an alternative goal (color). We then assign a \textit{user-intended} reward $\mathcal{R}'$ of high and low concept specificity (i.e. when the object is described by more or less attributes which makes object identification more specific) for TI vs. TR tasks (e.g. a TI reward would be ``go to \textit{any} goal'' whereas a TR reward would be ``go to the \textit{yellow} goal.'' For the third and fourth shift types (new distractor), we first generate the goal and a distractor in a similar manner, then randomly assign whether the distractor is TI (not in the way) or TR (in the way). For distractors, we similarly define their corresponding goals (e.g. a TI distractor would be ``go to the goal, ignore the distractor'' whereas a TR distractor would be ``go to the goal, ignore the \textit{red} distractor.'' For the last shift type (other), we generate the same goal instead in the bottom left corner. We do not explicitly define a test reward $\mathcal{R}'$ for the user---instead, we allow the user to \textit{implicitly} select this as the alternative option given no valid counterfactual is generated.

\textbf{Minigrid.}
We design a task, again defined as ``go to the $<$\texttt{goal}$>$'', with a multi-room compositional structure to explore how our approach scales with compositionality and long-horizon goals. We adapt the DoorKey environment from Minigrid \cite{minigrid} and create an environment composed of three sub-tasks (pick up a key, use the key to unlock a door, then navigate through the door to a goal). Success is defined by reaching the goal within 35 timesteps. Refer to Figure~\ref{fig:envs} for examples of sampled tasks.
% The state space is fully-observable and consists of RGB images of dimension 36$\times$36$\times$3. The action space is discrete of size 6 and allows for cardinal movements, picking up/dropping a key, and using the key to open a door.

% There are four possible objects: key, door, goal, and distractor. There is one visual concept (color) with 4 instantiations (red, green, blue, yellow) (other than lava, which has only orange and pink). 

\textit{Training task.}
We generate a task with randomly sampled key, door, and goal colors placed at fixed locations on the grid with no distractor. The reward $\mathcal{R}$ is the agent's distance from the goal. We create 10 demonstrations of length 35 by taking actions from the agent's starting location to the goal.

\textit{Test tasks.}
Our 5 potential distribution shifts are generated in a similar manner to above. For the first and second shift type (changed object), we first sample an object (key, door, goal), then randomly vary its concept instantiation (color). This is reflected in reward $\mathcal{R}'$ for TR shifts (e.g. ``go to the goal using the \textit{blue} key'' or ``go to the \textit{green}'' goal). For the third and fourth shift types (new distractor), we preserve the same key, door, and goal visual concepts from the train task, but additionally add a distractor of a random color (and assign it as TI or TR). For the last shift type, we sample an object and vary its location.

\textbf{VIMA.}
We design a visual manipulation task using VIMA \cite{jiang2022vima}, defined as ``put the $<$\texttt{object}$>$ on the $<$\texttt{goal}$>$''. In this domain, we wish to explore how more realistic environments with a larger concept set impact the accuracy and benefit from human feedback. VIMA consists of many objects and concept instantiations, and is therefore well suited for testing the impact of an increased concept set. We focus on a visual manipulation task where a robotic arm is tasked with picking up an object and placing it at a designated location on a tabletop surface. Success is defined as a successful pick and placed object within radius $\epsilon$ of the goal.

% States are fully-observable images of dimension 80$\times$80$\times$3 and represent a RGB fixed-camera topdown view of the scene. The action space is continuous of dimension 4 and consists of high-level pick-and-place actions parameterized by the pose of the end effector.

% There are 4 possible objects: the manipulated object, goal object, and two possible distractors. There are two visual concepts (shape and color) and 29 possible instantiations of shape and 81 instantiations of color, resulting in $(29\times81)^4\thickapprox 10^{13}$ possible scenes.

\textit{Training task.}
We generate a pick-and-place task where a robotic arm is tasked with picking up an object and placing it on the goal ($\mathcal{R}$: distance of the picked object from the goal). We place a randomly sampled goal object and its corresponding concepts in the bottom right corner of the table, then do the same for an alternately sampled manipulated object in the top left corner. There is no distractor. 

\textit{Test tasks.}
We generate 5 distribution shifts similar to above (Figure~\ref{fig:envs}). For the first and second shift type (changed object), we first randomly sample the object (e.g. manipulated or goal object), then vary randomly one of the properties (e.g. shape or color instantiation). TR $\mathcal{R}'$s would reflect this shift, e.g. ``put the object on the \textit{bowl}'' or ``put the \textit{green} object on the goal.'' For the third and fourth shifts (new distractor), we maintain the same object concepts from train, but additionally sample a distractor object and its concepts (and sample a corresponding TI or TR).

\textbf{Human experiments.} To explore whether counterfactuals help real human users identify task-irrelevant shifts, we conducted a between-subject in-person human experiment at the Massachusetts Institute of Technology. We recruited 20 subjects (55\% male, age 18-35). 24\% of participants attested to having a technical background, although only 10\% have worked with machine learning. We obtained IRB approval and all subject data was anonymized.

\textbf{Method.} The user study is comprised of two phases for each domain: familiarization and feedback. In the familiarization phase, we introduce the user to the task context, environment, concept space $\mathcal{S}_{\Phi}$ (object concept is defined in language, concept instantiation depicted visually akin to Figure~\ref{fig:features}), and an example of each type of distribution shift ($s_0'$ depicting object concept, distractor, or other shift). We then explain what a counterfactual demonstration is. In the feedback phase, we sample test tasks for each type of distribution shift (one each for 2D Navigation and Minigrid, 2 each for VIMA, amounting to 12 tasks per user). For the two shift types that can be TI or TR (shifted concept or distractor), we randomly assign half of the task specifications to be one type and half to be the other (in batches so that we test an equivalent number of total tasks). This is done to create comparisons between different ``user-desired rewards.'' For full examples across all domains, see Figure~\ref{fig:envs}.

\textbf{Baseline condition.} For each test task, we first show the user the failed behaviour and ask them to identify, of the defined $\mathcal{S}_{\Phi}$, which 1) concepts(s) and 2) instantiation(s) they believe caused the failure. We then ask whether these concept(s) should matter to the specified reward $\mathcal{R}'$. We highlight that $\mathcal{S}_{\Phi}$ may not contain all the visual concepts relevant to failure (i.e. something other than object concept or distractor may be the error, e.g. location). We allow responses of none, one, or multiple visual concepts (with none corresponding to ``the cause of failure is not in $\mathcal{S}_{\Phi}$''). 

\textbf{Experimental condition.}
Next, we show a counterfactual demonstration generated via an oracle demonstrator providing a desired demonstration. Note, in a real world deployment scenario, this demonstration would be collected from the user. We then repeat the same process as the baseline. Altogether, we received 240 user responses to 12 tasks.

%% file: 5-results.tex
\section{Empirical evaluation.}
\label{sec:results}

\begin{figure*}
    \centering
    \includegraphics[width=.8\textwidth]{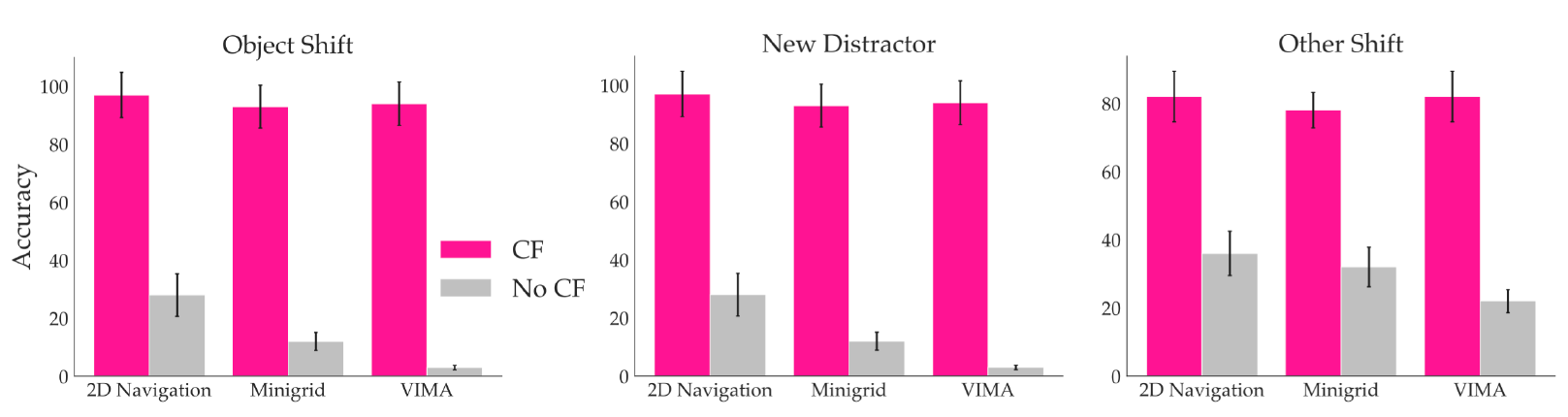}
    \caption{User accuracy on identifying the correct shifted concept. ``No CF'' represents our baseline condition where the user only had behaviour, while ``CF'' represents our experimental group where the user also had counterfactual demonstrations. Error bars denote mean standard error.}
    \label{fig:accuracy}
\end{figure*}

Our approach aims to improve the efficiency of policy adaptation of task-irrelevant shifted concepts in $\Delta\phi$ while minimizing user effort required to generate demonstrations---\textit{does this work in practice with real humans?}.

\textbf{Q0 (sanity check): Can users select $\Delta\phi$ from scratch?}
Before investigating our main hypothesis, which is whether counterfactual explanations can help users infer the true shifted $\Delta\phi$, we conducted a pilot user study in VIMA (N=5) to determine whether participants could accurately specify task-relevant concepts with no machine guidance, i.e. can users correctly identify the correct policy failure mode without a provided explanation? We followed the same user study protocol described above, but asked for responses with no counterfactuals presented. A response was counted as correct if the true task-relevant concepts were selected.

\textbf{Results.} We pooled responses across 5 randomly sampled tasks per user, and found that none of the five respondents were able to correctly identify the correct relevant subset above random chance. For detailed discussion regarding specific user failures, see \ref{supp:user-study-details}. This confirms the intuition from related work \cite{bobu2018learning,ilyas2019adversarial} that users generally struggle to \textit{exhaustively} specify all task-relevant concepts that agents also find relevant due to the \textit{black-box} and sometimes arbitrary nature of policy failures.

\textbf{Q1: Do counterfactuals help the user better infer $\Delta\phi$?}
Now that we have performed a sanity check motivating the need for an explanatory aide in the user-feedback phase, we begin our analysis of the benefit of interactive human diagnosis of distribution shift. We first ask whether our proposed explainability method of generating counterfactual demonstrations results in more accurate human inference of $\Delta\phi$ when compared to agent behaviour alone, i.e. do counterfactuals help users identify the true shifted concept(s)?

We compare the accuracy of user-provided responses to the true shifted concept(s) $\Delta\phi$ from behaviour alone (our baseline condition) vs. with counterfactual demonstrations (our experimental condition). For a response to be counted as correct for a concept shift, the user must have identified both the correct 1) object and 2) concept within their response. For identifying new distractors, a correct response must include all visual concepts of the distractor (e.g. color and also shape if present). For ``other'' shifts with no counterfactuals, users who responded with ``none'' were correct. 

\textbf{Results.} As highlighted in Figure~\ref{fig:accuracy}, users who received counterfactuals were \textbf{significantly more accurate at identifying the correct concept shift} compared to users with behaviour alone ($p$ $<$ .001 across all domains using a Chi Squared test with 1 degree of freedom $\tilde{\chi}^2(1)$.). Unsurprisingly, this effect is the largest on VIMA, which contains by far the largest concept space. This result confirms our hypothesis that we can leverage user feedback for more accurate augmentation in more complex domains. Lastly, these results highlight the value of leveraging counterfactuals as an interpretability tool for sequential decision-making.

\textbf{Q2: Does our approach result in more efficient finetuning?}
Now that we have reinforced the ability of counterfactual demonstrations to provide more accurate user inference of $\Delta\phi$, we ask: can we successfully leverage additional feedback regarding \textit{task-irrelevant visual concepts} to more efficiently perform augmentation and finetuning?

To answer this, we turn our empirical analysis to the two types of distribution shift that possess task-irrelevant visual concepts: object concept and distractor shifts where the desired reward is ``broad'' and may contain ambiguous aspects that should not impact actions. For each test task, we assess performance of the finetuned policy on 10 sampled tasks from $\mathcal{R}'$. We make the following comparisons:

\textbf{No human feedback (NH-random)}: We attempt random augmentation of all possible concepts in the demonstration. While generating augmented demonstrations is less costly relative to collecting them from humans, finetuning on all permutations (without being confident that it would result in the right solution) is still undesirable.
%given higher-complexity environments with a large concept set.

\textbf{Baseline human (Baseline H)}: Our baseline human group, where subjects identified visual concepts from the agent's behaviour alone. We augment the correct concept at the same rate of correct responses for each domain, and randomly otherwise.

\textbf{Counterfactual human (CF H)}: Similar to above, but with responses from our experimental group.

\textbf{Perfect feedback (Oracle)}: For completion, we also include a comparison to the maximum performance gain possible if we always augmented the correct concept.

\begin{figure*}
    \centering
    \includegraphics[width=.8\textwidth]{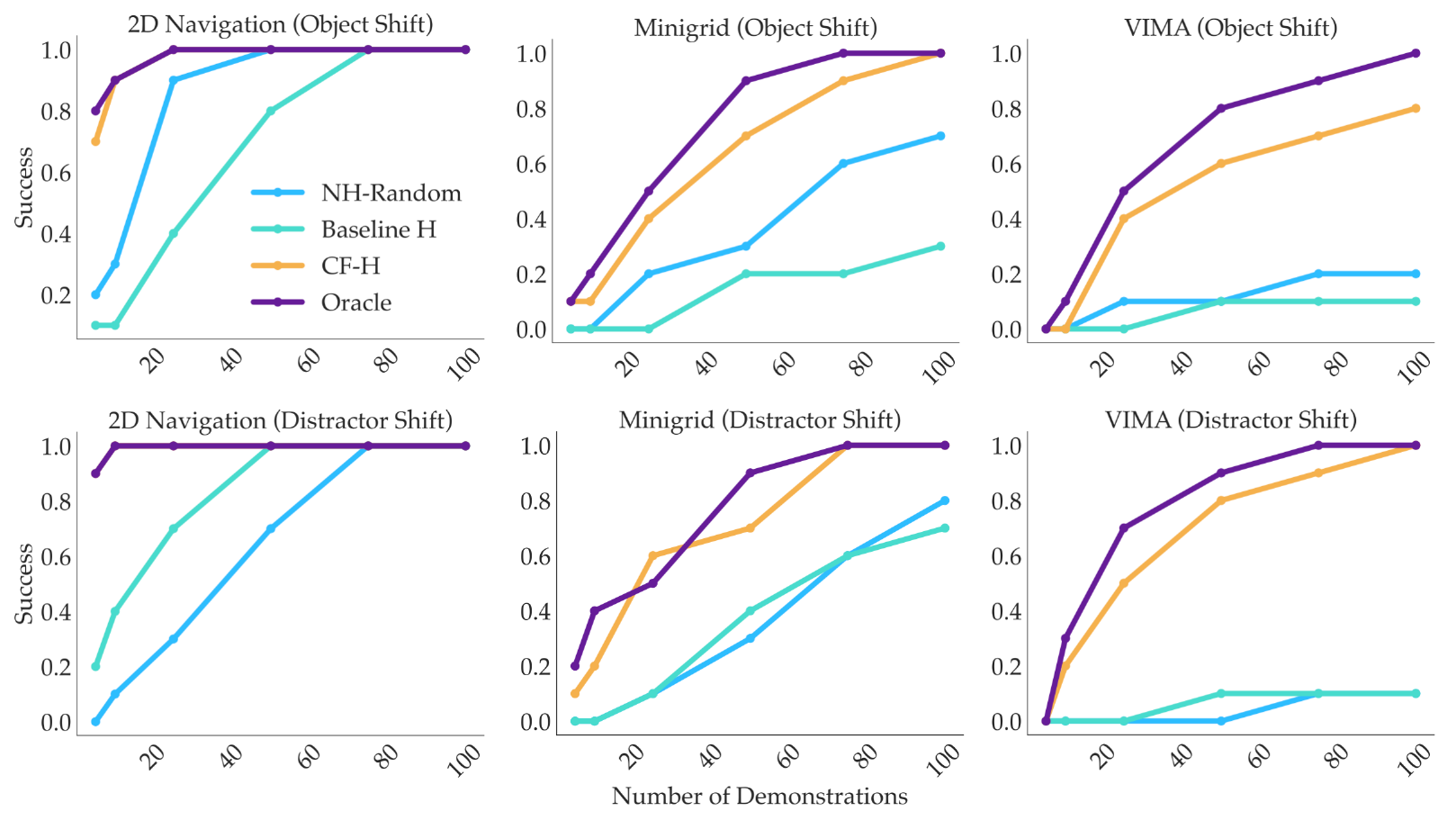}
    \caption{Finetuning results on task-irrelevant shifts across all domains.}
    \label{fig:plots}
\end{figure*}

\textbf{Results.} As shown in Figure~\ref{fig:plots}, more accurate feedback directly results in \textbf{more efficient policy performance on $R'$ after targeted data augmentation}. These results illustrate an exciting future direction for interactive methods that query for concept-level human invariances on the test task for more efficient learning.

\textbf{Q3: Does our approach align to human desired rewards?}\newline
The previous experiments considered questions related to a user's ability to infer $\Delta\phi$ and their behaviour-invariance---properties that inform of distribution shifts caused by the \textit{state}. However, we now evaluate whether our approach also allows users to identify \textit{reward} shifts i.e. can users appropriately identify task-irrelevant concepts for different $R'$s?

\textbf{Results.} The answer to this question lies in our random assignment of TI and TR rewards to different object and distractor shift tasks. For TR assigned tasks (e.g. go to the \textit{blue} object), human subjects correctly identified that the proposed counterfactual (e.g. \textit{red} object) was \textit{not valid for their desired reward} 79\% of the time across all domains. In a pairwise comparison between responses to those TR tasks and their corresponding TI tasks, we find a significant difference in respondent answers ($p$ $<$ .001 across all domains using a Wilcoxon signed-rank test). This points to evidence that users \textbf{do not indiscriminately view all concepts similarly} but rather \textbf{ground them to their specific $R'$}. This is an exciting finding, for it confirms that humans possess preferences that may be \textit{under-specified} by the task definition that we can query to perform user-guided adaptation.

%% file: 6-related_work.tex
\section{Related Work}
\label{sec:related_work}

\textbf{Human-in-the-loop.} Existing frameworks to interactively query humans for data \cite{abel2017agent,zhang2019leveraging}, like TAMER \citep{knox2008tamer} and COACH \citep{macglashan2017interactive} use human feedback to train policies, but are restricted to binary or scalar labeled rewards. A different line of work seeks to learn tasks using human preferences, oftentimes asking them to compare or rank trajectory snippets \citep{christiano2017deep,brown2020better}. Yet another direction focuses on how to perform active learning from human teachers, where the emphasis is on generating actions or queries that are maximally informative for the human to label \citep{bobu2022learning,chao2010transparent}. The challenge with these approaches is that the feedback asked of the human is often overfit to specific failures or desired data points, and rarely scale well relative to human effort \cite{bobu2023aligning}. Our approach scales exponentially with the number of visual aspects present in the scene---given an ability to modify them if identified.

\textbf{Counterfactual explanations.}
Ensuring that end-to-end agents are intelligible to various concerns encompassing ethical, legal, safety, or usability viewpoints is a key focus of counterfactual explanations \citep{garcia2015comprehensive}. While commonly found in image classification \cite{goyal2019counterfactual,vermeire2022explainable}, counterfactuals have been scantily explored in sequential decision-making. The nearest neighbor to our work is \cite{olson2019counterfactual}, who use a GAN to generate counterfactual states directly. However, the use of counterfactuals to solicit useful human feedback for adaptation has been under-explored, and to the best of our understanding, never done previously for robotic tasks.

\textbf{Visual Augmentation.} Recent works conducted in parallel to ours explore the feasibility of leveraging visual augmentation via image editing to train more generalizable policies, and optimistically suggest that augmentation of visual concepts alone buys quite a lot in real-world robotics settings \cite{chen2023genaug,mandi2022cacti,yu2023scaling,chen2021understanding}. While these works focus exclusively on technical methods to scale up visual augmentation (often by using image editing techniques to augment designer-specified concepts), they do not address the underlying question of identifying specific \textit{user-informed} task-irrelevant concepts that we do in our work. We concur with the suggestions offered in these works for scaling beyond visual data, such as mixing in simulation data for capturing motion, exploring multi-modal data generation, etc., and are excited about the complementary nature of these two lines of work.

%% file: 7-discussion.tex
\section{Discussion}
\label{sec:discussion}
%\pulkit{SUmmary is not needed in the Discussion -- its like the abstract which is already there in the paper :)}
%\textbf{Summary.}
%In this work, we proposed a framework to query for task-irrelevant concepts of the state and reward causing distribution shift to perform targeted data augmentation for efficient finetuning. We successfully used counterfactual demonstrations to identify and augment TI concepts. Experiments on three different domains with real human subjects confirmed the data efficiency of our findings. Overall, our work conveys the optimistic message that exploring methods for communication between humans and agents at the task concept level enables faster, user-guided adaptation. 

\textbf{Limitations.}
Our work relies heavily on Assumptions 1 and 2 of access to an abstracted state space and image editor $f$ for a desired scene.
This abstraction may be difficult to obtain in many situations in the wild, where scenes could be infinitely cluttered with many objects. However, the exciting developments emerging from the field of computer vision enabling fast, realistic image editing grounded in human-defined abstractions \cite{mittal2023orbit,gal2022image,kawar2022imagic} is an encouraging sign. Moreover, while we define concept invariance to encompass either one specific instantiation (e.g. \textit{blue}) or all instantiations---users may possess more complex delineations (e.g. \textit{blue} and \textit{green}, but not \textit{red}). Future work can also explore how different visualization interfaces helps users give even more detailed feedback regarding their preferred aspects.

More importantly, our minimum edit counterfactual solution is solved in this work via privileged access to object ordering, then a brute force search over all their instantiations, which may not be practical. Future work can explore how continuous relaxations may be applied to the problem so that a method like gradient descent can be employed to generate counterfactuals. We are excited for the following two directions of inquiry: (1) leveraging human heuristics and (2) learning a prior over counterfactuals. For the first, one possibility is to use domain knowledge to design heuristics, e.g. first identify whether the presence of an object impacts behaviour prior to performing a search over each possible visual instantiation (an idea that we implemented in our work). We envision that other heuristics can be additionally explored to constrain the search space. For the second, we could also learn a counterfactual priors model to automatically extract these heuristics using data from similar already-seen tasks. This could be done by learning which concepts correspond to different types of pretraining tasks, shifting the burden of structuring the search space from deployment time to training–a reasonable path forward given our goal of fast adaptation to the end user.

%% file: 8-acknowledgements.tex
\section{Acknowledgements}
We thank Abhishek Gupta, Anurag Ajay, Jacob Huh, and the members of the Improbable AI Lab and Interactive Robotics Group for helpful feedback and discussions. We additionally thank Yunfan Jiang for technical help with setting up and debugging VIMA. 

We are grateful to MIT Supercloud and the Lincoln Laboratory Supercomputing Center for providing HPC resources. Andi Peng is supported by the NSF Graduate Research Fellowship and Open Philanthropy, and Andreea Bobu by the Apple AI/ML Fellowship. This research was also partly sponsored by Hyundai Motor Corporation, the MIT-IBM Watson AI Lab, and the NSF Institute for Artificial Intelligence and Fundamental Interactions (IAIFI).

%% file: 9-appendix.tex
\appendix
\onecolumn
\section{Appendix}

\begin{figure*}
    \centering
    \includegraphics[width=.95\textwidth]{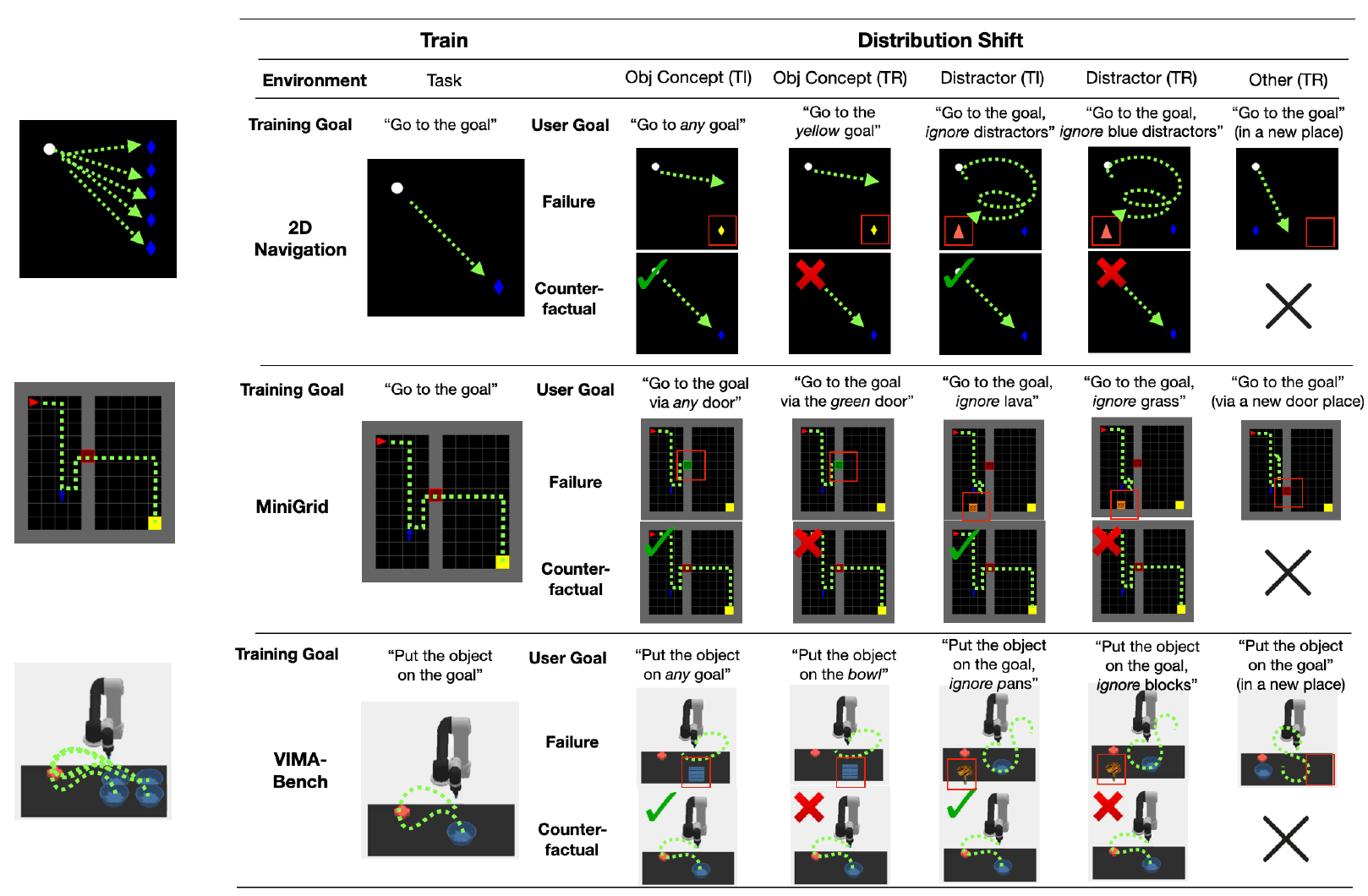}
    \caption{Example training tasks and sampled distribution shifts for test tasks for all domains.}
    \label{fig:envs}
\end{figure*}

\subsection{Additional Environment Details}
\label{supp:implementation-details}
% In practice, we experiment with environments where $k\leq1$ and possible visual concepts include object color and shape. The number of objects is fixed and $n\leq4$.

\textbf{2D Navigation.} States are fully-observable RGB images of dimension 36$\times$36$\times$3. The action space is continuous and represents the $(x,y)$ distance that the agent can move in 2D space. The environment was created to test a simple visual domain while preserving a continuous action space.
There are two possible objects: goal and distractor. There is one visual concept (color) with 4 instantiations (red, green, blue, yellow).

\textbf{Minigrid.} The state space is fully-observable and consists of RGB images of dimension 36$\times$36$\times$3. The action space is discrete of size 6 and allows for cardinal movements, picking up/dropping a key, and using the key to open a door.
There are four possible objects: key, door, goal, and distractor. There is one visual concept (color) with 4 instantiations (red, green, blue, yellow) (other than lava, which has only orange and pink). 

\textbf{VIMA.} States are fully-observable images of dimension 80$\times$80$\times$3 and represent a RGB fixed-camera topdown view of the scene. The action space is continuous of dimension 4 and consists of high-level pick-and-place actions parameterized by the pose of the end effector.
There are 4 possible objects: the manipulated object, goal object, and two possible distractors. There are two visual concepts (shape and color) and 29 possible instantiations of shape and 81 instantiations of color.
% , resulting in $(29\times81)^4\thickapprox 10^{13}$ possible scenes.

\subsection{Pilot User Study Details}
\label{supp:user-study-details}

In the user debrief from the pilot user study motivating counterfactuals, we found two main user failure modes:

\textbf{1. Over-specification.}
2 out of 5 users said they included more task-relevant concepts than required because they were not sure how these abstract concepts would actually impact behaviour. For example, if the task was “pick up the flower and put it on the pan” (ground truth concept set being target shape and goal shape), these users additionally included non-relevant concepts like target color and goal color. Hence, in the absence of explanations grounding the abstract concept space to behaviour in the environment, these users felt they could not definitively tell which concepts were relevant vs. not, therefore opting to include everything.

\textbf{2. Under-specification}
The other 3 users under-specified task-relevant concepts by struggling to imagine alternative future scenes that the robot could be deployed in. For example, if the task was “pick up the object and put it on the blue pan” (ground truth concept set being goal shape and goal color), these users only selected goal color as a task-relevant concept without realizing that the goal itself could take on many different forms. This is due to humans automatically thinking of concepts through their own interpretation (e.g. the goal is a pan right now and so it must always be a pan) without considering other alternative realities (e.g. goals can be other objects). 

Counterfactuals helped explicitly highlight the axes of variation that exist in an environment, helping the user understand that their current interpretation of the concept is different than all instantiations that concept could be in the scene.